\documentclass[11pt]{article}
\usepackage{hyphenat}

\usepackage[preprint]{acl}

\usepackage{times}
\usepackage{latexsym}

\usepackage[T1]{fontenc}

\usepackage[utf8]{inputenc}

\usepackage{microtype}

\usepackage{inconsolata}

\usepackage{graphicx}
\usepackage{paralist}
\usepackage{booktabs}
\usepackage{hyperref}

\title{Re-defining Humor Data Objects for AI Humor Research}

\author{
 \textbf{Anna Arnett},
 \textbf{Bang Nguyen},
 \textbf{Meng Jiang}
\\
 Department of Computer Science and Engineering, University of Notre Dame
}

\begin{document}
\maketitle
\begin{abstract}
In most existing AI humor research, humor was treated as either ``present'' or ``not present.'' We explore the concept of humor as a social interaction with context and explanations. During this project, we defined a \emph{humor reasoning} data object and developed a way to prompt LLMs to generate an explanation of humor effective for general population. We iterated from an earlier prompt to an improved prompt, found that the later version reduced important errors, and then scaled generation to a large number of data objects which have the potential to enable data synthesis and data augmentation for AI humor research. Our main takeaway is that better prompting of an LLM improves humor explanation quality, especially by handling missing context, multi-modality, and transcript issues more carefully. These results establish a strong foundation for future work on AI understanding of humor as social behavior. All code and data are available at: \url{https://github.com/anna-arnett/ai-humor/}
\end{abstract}

\section{Introduction}

Existing work on evaluating and studying the AI's understanding of humor represented humor as a textual object; thus, humor detection was a binary classification task, and humor datasets were binary labels on texts \cite{weller2019humor,annamoradnejad2020colbert}. However, in psychological research, humor is defined as \emph{social behavior} \cite{martin2006psychology}, which is a fundamental tool for social interaction, communication, relationship building, and the establishment of group norms. In this work, we make multi-fold contributions: we think outside the binary-classification box, re-define the ``humor'' data object from the perspectives of social roles involved in the interaction, build a novel dataset for multiple new humor-related tasks, and comprehensively study AI's ability of understanding humor. Our preliminary work focused on two concrete milestones: outlining the ideal structure of a humor data point, and prompting an LLM to generate this data point. The next step will be generating large-scale humor reasoning data objects, and validating their usefulness in data synthesis and data augmentation for building humor-related AI models. For example, by prompting to change key concept in a humor reasoning data object, one can create a humor alternative or a non-humor data point where the modification destroys the reasoning. Training such synthetic data may improve the performance of humor detection. 

\section{Related Work}


\subsection{Humor Capabilities in AI and LLMs}
The literature has provided a strong motivation for investigating and enabling humor capacity in AI systems. While humor understanding tasks test an AI's ability to recognize and explain humor from textual descriptions, generation tasks investigate the quality and social appropriateness of AI content \cite{10.1145/3778357}. More recent advances have integrated additional modalities beyond text into these applications, such as images \cite{CAO2025100807}, video, and speech recordings \cite{10707160}. 

AI humor has implications for various domains. Studies have demonstrated the potential adoption of AI humor for screenwriting and news writing \cite{toplyn-amir-2025-ai, mirowski-etal-2025-theate}. When equipped with a humorous personality, AI systems are found to be likable \cite{10.1145/3383652.3423915, CAO2025100807} and trustworthy \cite{marone-2025-algorithm, kolomaznik2024role}, which in turn provides meaningful implications for their application as mental health assistants \cite{10.1145/3383652.3423915}. Recognizing these diverse applications, we are interested in defining a well-informed data object that facilitates more systematic AI humor research.

\subsection{Humor Understanding and Generation}
Previous work has found that word embeddings can capture humor knowledge in their representations and reflect existing social humor theories \cite{skalicky-attardo-2025-testing, 10.1145/3606039.3613102}. Training-based approaches adopt RNN-based and Transformer-based architectures on domain- and language-specific datasets \cite{10.1145/3689062.3695939,LI2020102290}. The training of multi-modal humor recognition and generation models typically happens first through the individual training of unimodal feature extraction models, which are then combined using feature fusion strategies \cite{singh2024exploring, 10707160}. More recent work shifts focus toward LLMs and their in-context learning abilities, integrating social and psychological theories through prompting to improve the performance of AI in humor understanding and generation \cite{baluja-2025-text, sarrof-2025-homophonic, lee-etal-2025-pragmatic, joshi-2025-evaluating}. Across these methodologies, humor is typically defined as a single textual object. Recognizing the social nature of humor \cite{martin2006psychology}, we propose a revised data object for AI humor research; in this object, we consider not only the humorous attempt or joke with a binary label, but also the associated social context, the audience's reaction, and potential recovery attempts for failed humor.

\section{The ``Humor'' Data Object}

Humor happens between two social roles: teller and receiver. In a social dialogue or context, the teller makes a humorous attempt and expects the receiver's reaction. The receiver may feel humorous, which may be observed as laughter, or not humorous. This signal has been numerously collected in existing humor datasets. The receiver actually can explain his/her reaction: it is a quick yet \emph{step-by-step reasoning} process which may involve relevant cultural knowledge or commonsense; however, this is always missing in existing studies.

Reasoning and reaction is not the end. If the reaction is negative (i.e., not humorous), the teller usually makes a \emph{recovery} response after the humor failure. Different strategies can be applied in a reasonable recovery, such as deflection, clarification, and apology.

In this work, we define a humor data object as a quintuple $(C, X; Y, R; Z)$:
\begin{compactitem}
    \item $C$ is social dialogue or context;
    \item $X$ is teller's humorous attempt;
    \item $Y$ is receiver's reaction (binary): $1$ means feeling humorous; $0$ otherwise;
    \item $R$ is receiver's explanation to his/her positive or negative reaction: it explains why humor succeeds or fails;
    \item $Z$ is teller's recovery response after humor failure ($Y=0$); it's `n/a' if $Y=1$.
\end{compactitem}

Our preliminary work focused primarily on R, the receiver explanation. Looking further into Z (recovery after failure) remains future work. 

\section{Dataset Construction}

\subsection{Context, Attempt, and Reaction}

We started from existing datasets that contain observable reactions, especially SMILE (TED+sitcoms) \cite{chen2023smile}. For each instance, we represented the data point using the local textual context $C$, the candidate humorous utterance (or "punchline") $X$, and the observed audience reaction $Y$. Specifically, $X$ can be the remark marked with receiver's laughter (or no laughter); $C$ can be preceding sentences from the same speaker and/or dialogue. 

Our main goal at this stage was to create inputs to feed into an LLM and generate a humor explanation. We weren't focused on annotating every possible aspect of the source data. 

Because many source examples originated from multimodal settings like TED Talks or sitcoms, the transcript was often missing important information about the context of the joke or what was shown to the audience at the time the punchline was made. This limitation became very important later in the project, because some examples depended on a visual image or cue that was not recoverable from the text alone. As a result, transcript quality and sufficiency of concepts became central concerns to generating humor explanations. 

A JSON schema for the examples we used is shown below:
\begin{verbatim}
  {
    "id": "414",
    "source": "TED",
    "C_text": [...],
    "X_text":  [...],
    "X_speaker": [...],
    "Y": 1,
    "model_output": {
        "text_support_label": [...],
        "data_quality_flags": [...],
        "reasoning_steps": [...],
        "flow_chart": [...],
        "final_explanation": [...],
        "hallucination_flag": 0
    }
}  
\end{verbatim}

\subsection{Explanation Generation}

\subsubsection{Initial Prompt}
Our first prompt was an initial attempt to generate a \emph{structured explanation} of why a joke succeeded in context. Given the preceding context $C$, the punchline $X$, the speaker, the observed reaction $Y$, the model was asked to return a JSON object containing step-by-step reasoning, a causal flow chart, a one-sentence final explanation, and a hallucination flag. 
This design is motivated by prior work in psychology related to humor. Humor theory suggests that the process of finding something humorous happens in rapid but structured mental steps \cite{suls1972twostage, martin2006psychology}. By requiring explicit intermediate steps in our process, we aim to externalize these otherwise implicit processes to make the reasoning behind humor observable.

In many text-sufficient cases, the first prompt output useful explanations. However, we manually reviewed the outputs, and the first prompt also had some limitations. We double-checked 31 generated explanations one-by-one and judged whether each one matched our opinion of the data point. Under this review, 17 of the 31 examples were acceptable, while 14 were not. Seven of the errors were due to \underline{incomplete context}, two errors were due to \underline{named entities not being properly explained}, and five errors were due to \underline{multimodality}. In particular, the prompt often tried to hallucinate a full explanation even when the transcript did not provide enough context for one. This motivated us to develop a revised prompt that more explicitly evaluated whether the text was sufficient on its own. 

\subsubsection{Prompt Revision}
We devised a new prompt in response to the issues we identified with Prompt 1. These issues made us think that the model needed to assess whether the transcript itself was sufficient to support the joke explanation. 

To address this, the second prompt extended the task in two ways. First, it required the LLM to specifically assess whether the humor reaction made sense from the text alone. Second, it asked the model to flag any possible data quality issues such as missing context, alignment problems, or multimodal dependence. This prompt better distinguished between the cases where the humor could be explained from the text and the cases that it couldn't, rather than hallucinating an explanation no matter what. 

We tested the revised prompt on the same 31 manually reviewed examples used earlier. Under this second prompt, 27 of the 31 outputs were correctly labeled. The remaining four errors were also narrower and more specific than before: one case failed because the model did not detect that the dataset was missing context, and three cases failed because the model did not detect likely multimodal dependence. Compared with the first prompt, this was a big improvement.

\subsubsection{Scaling Generation}
After revising the prompt, we used this new prompt to scale up our dataset beyond the initial 31 examples. We started with a small pilot of examples to generate and manually inspected the output to check if it was valid JSON, if the reasoning remained solid based on the transcript, and if the model did not hallucinate humor explanation. The pilot outputs appeared promising enough to proceed to a larger run of generating outputs.

Using this same generation pipeline, we produced a dataset of 307 explanation examples. As a result, the generated dataset not only contained explanations of why humor succeeded, but also some metadata about if the transcript itself was sufficient to support that explanation. We found that this dataset serves two purposes. First, it provides a resource to study explanation generation in humor understanding. Second, it taught us that a well-designed prompt can be used as a scalable technique for creating humor data points, even when the transcript is flawed. This dataset forms a solid foundation for future work on humor data synthesis, humor data augmentation, and humor understanding model development.

\textbf{Implementation details.}
All explanations were generated using the same base LLM checkpoint: \texttt{[gpt-4o-mini]}. We implemented two prompt versions. For manual evaluation, we applied both prompts to the same 31 examples and judged each output as acceptable or unacceptable based on whether it matched our interpretation of the transcript. To generate responses on a larger scale, we first ran a small pilot and then used the revised prompt to generate 307 explanation examples. The full text of Prompt 1 and Prompt 2 is included in the appendix for reproducability.

\section{Humor Reasoning Evaluation}

Given $C$, $X$, and $Y$, the goal of Task 2 is to generate an explanation $R$ for why humor succeeds or fails. This task became our main focus in the preliminary work of this project. In particular, we developed a prompt-based method for generating $R$. The explanation/reasoning task is evaluated according to the following criteria:
\begin{compactitem}
    \item Explanation relies only on $C$ and $X$, or explicitly stated common sense assumptions (Groundedness);
    \item Explanation identifies why humor works or fails (Completeness);
    \item Explanation identifies the true humor mechanism;
    \item Explanation does not introduce unsupported facts (yes/no).
\end{compactitem}
These criteria are based on humor theory and social cognition research \cite{suls1972twostage, martin2006psychology}. In our preliminary work, we mostly evaluated through manual review on a small subset of examples. This manual analysis was especially useful for identifying failure cases. In future work, this task could also be evaluated with an LLM-as-a-judge approach after validating the rubric against our human judgments.

\section{Results and Lessons Learned}

\subsection{Results}

Our work culminated into two main milestones. First, we defined a structured humor reasoning data point. This data point represents humor as a social interaction, not just a binary flag of presence or non-presence. The data point involves context, a humorous attempt, a reaction, and an explanation. Second, we created a method using LLM prompts to generate the explanation of humor at scale. 

Our first prompt already produced acceptable outputs in a number of cases, especially when the transcript was very sufficient in order to understand the joke. On a manually reviewed set of 31 examples, 17 outputs were judged to be correct, while 14 were not. The failures were informative: 7 were mainly due to incomplete context, 2 were due to named entities or references not being sufficiently explained, and 5 were due to multimodal issues, where the punchline of the joke relied on images/gestures or a general delivery not available in the transcript. 

\begin{table}[h]
\centering
\begin{tabular}{lcc}
\toprule
& Correct / Acceptable & Incorrect \\
\midrule
Prompt 1 & 17 / 31 & 14 / 31 \\
Prompt 2 & 27 / 31 & 4 / 31 \\
\bottomrule
\end{tabular}
\caption{Manual evaluation results on the same 31 explanation examples.}
\end{table}

\subsection{Detailed Failure Case Studies from Prompt 1}
To better understand these errors, we grouped the incorrect outputs from Prompt 1 into a small set of recurring failure categories. 

\paragraph{Incomplete context.} 

In this category, the transcript provides a local setup and a marked laughter point, but it excludes some key background information that is needed to understand the joke.

A representative incomplete context example is shown below: 

\begin{quote}
    \textbf{context $C$:} ``\emph{Nagin Cox: so one day you come in at the next day minutes later at the next day minutes later at the next day at, so you keep moving minutes every day until soon you're coming to work in the middle of the night the middle of the earth night, right so you can imagine how confusing that is}''

    \textbf{humor attempt $X$:} ``\emph{hence the mars watch (audience laughs)}''
\end{quote}

As we can see, the transcript itself does not provide enough context to to explain why this phrase is funny or what prior setup the audience is expected to recognize. 

Under Prompt 1, the model often hallucinated a complete explanation even when there was not enough context. This case motivated the revised prompt's emphasis on \texttt{text\_support\_label} and explicit detection of insufficient context. 

\paragraph{Case 2: Named entities not explained.}

In this category, the transcript might contain enough information to understand the joke, but the model fails to explain a real-world reference in its explanation. 

For example:

\begin{quote}
    \textbf{context $C$:} ``\emph{a few times a year they tell their engineers to go for the next hours and work on anything you want as long as it's not part of your regular job work on anything you want engineers use this time to come up with a cool patch for code come up with an elegant hack being australians everybody has a beer they call them fedex days}''

    \textbf{humor attempt $X$:} ``\emph{it's a huge trademark violation but it's pretty clever}''
\end{quote}

In this example, the humor makes a lot more sense when the model explains how FedEx is associated with overnight delivery. Prompt 1, instead of explaining this, described the line as merely ``clever,`` which is not thorough enough to make sure that any reader could understand. 

\paragraph{Case 3: Multimodal dependence.}
In this category, the laughter appears to depend on an image, gesture, facial expression, timing cue, or some other non-textual event that is not recoverable from the transcript alone. For example:

\begin{quote}
    \textbf{context $C$:} ``Sendhil Mullainathan: \emph{i want you to do it fast and say it out loud with me and do it quickly, i'll make the first one easy for you, ready black}''

    \textbf{humor attempt $X$:} ``\emph{red (audience laughs)}''
\end{quote}

From the transcript alone, the speaker is clearly leading the audience through some kind of rapid response task. However, it is not clear why the answer ``red`` is funny. It is plausible here that the humor depends on an unseen visual stimulus.

This was difficult to decipher from the text alone. However, Prompt 1 tended to, again, hallucinate an explanation regardless, leading to errors. 

\paragraph{Case 4: Unreliable transcript/alignment error.}
Our input dataset was imperfect, and some examples appear to contain mistranscribed wording, typos, or omitted words. In these cases, the model would try to produce a coherent explanation even though the source text was semantically broken. 

 An example of this is the following:

 \begin{quote}
     \textbf{context $C$:} ``Sheryl WuDunn: \emph{they have a hut that has no electricity no running water no wristwatch no bicycle, and they share this great splendor with a very large pig, dai was in sixth grade when her parents said we're going to pull you out of school because the school fees are too much for us, you're going to be spending the rest of your life in the rice paddies, why would we waste this money on you this is what happens to girls in remote areas, turns out that dai was the best pupil in her grade, she still made the two hour trek to the schoolhouse and tried to catch every little bit of information that seeped out of the doors, we wrote about her in the new york times}''

     \textbf{humor attempt $X$:} ``\emph{we got a flood of donations mostly checks because new york times readers are very generous in tiny amounts but then we got a money transfer for really nice guy (audience laughs)}''
 \end{quote}

In this example, the transcript is unreliable in 2 ways. First, the phrase ``for really nice guy`` is semantically incomplete and suggests missing words. Second, the laughter marker itself appears to be misaligned: the audience reaction was actually attached to the earlier phrase ``new york times readers are very generous in tiny amounts`` rather than to the final clause. This means that the model is trying to explain laughter using a punchline that it both broken and attached to the wrong place.

Prompt 1 often failed on examples like this because it tried to interpret the incorrect text as if it were a coherent punchline. Cases like this motivated the revised prompt's emphasis on transcript quality and wording/alignment issues.

These case studies clarified for us that many errors were not random. They also guided the design of Prompt 2, which explicitly asked the model to judge whether the text itself was sufficient and to flag likely data quality problems.

The revised prompt noticeably improved this behavior. On the same set of 31 examples, 27 outputs were judged to be correct. The remaining 4 errors were narrower than before: 1 case involved missing context that the model did not explicitly detect, and 3 involved likely multimodal dependence that the model still failed to flag. Compared with the first prompt, the revised version was more cautious and better at distinguishing between cases where the humor mechanism was recoverable from text and cases where the transcript was insufficient. This improvement was the clearest technical result.

Using the improved prompt, we generated a dataset of 307 explanation examples. In particular, the revised prompt more often avoided forcing an explanation when the source transcript appeared unreliable. 

We also explored automatic prompt optimization using DSPy/GEPA. This experiment was useful as a foray into different methodologies, but it did not become a central result. In our setup, \emph{automatic prompt optimization did not improve performance beyond adjusting the prompt manually}. This suggested that the main gains came from targeted prompt design informed by human error analysis, rather than from automatic optimization under the current evaluation setup.

\subsection{Lessons Learned}

Several lessons emerged from this work.

First, transcript sufficiency is one of the most important issues in humor explanation generation. Many humor instances originate from multimodal settings such as TED talks or sitcom scenes, and the transcript alone may omit visual cues, timing, delivery, or earlier setup that contributed to audience laughter. As a result, a good explanation system must do more than just generate a humor explanation: it must also assess if the available text actually supports the explanation.

Second, prompt design matters substantially. The shift from the first prompt to the revised prompt did not depend on changing the base model. Instead, it came from identifying failures through manual review and then explicitly incorporating those concerns into the prompt. 

Third, manual evaluation was essential. The most useful improvements came from reading examples one by one and identifying where the prompt failed. This made it possible to distinguish between different classes of errors, such as missing context, unclear references, multimodal dependence, and transcript noise.

Finally, our preliminary work showed that it is feasible to generate structured humor reasoning data points with a large language model at a useful scale. The resulting 307-example dataset is not a final gold-standard resource, but it demonstrates that explanation generation for humor can be done in a structured way. This provides a strong base for future work on evaluation, dataset expansion, and downstream uses of humor reasoning data.

\section{Conclusions and Future Work}

In this project, we introduced a \emph{humor reasoning} data object and developed a prompting approach for generating humor explanations that are accessible to a general audience. Through iterative prompt refinement, we found that the improved prompt reduced key errors and produced higher-quality explanations, particularly in cases involving missing context, multimodal information, and imperfect transcript data. We then applied this approach at scale to generate a large set of humor reasoning data objects, demonstrating its potential value for data synthesis and data augmentation in AI humor research. Overall, our findings suggest that careful prompt design can meaningfully improve the reliability and usefulness of LLM-generated humor explanations.

Several natural next steps follow from this foundation. First, the generated explanation dataset can be expanded and evaluated more systematically. Although the current 307-example file appears promising based on our first inspections, we should evaluate it in a more detailed way in the future. In particular, it would be valuable to measure explanation quality separately for text-sufficient cases, missing-context cases, multimodal cases, and transcription-error cases. This would provide a clearer picture of where the output is reliable and where transcript limitations remain a major obstacle.

Second, future work could investigate whether this dataset improves other humor-related language tasks. One especially interesting direction is to see if an LLM can generate better jokes when it is given this explanation dataset to train on.

Third, the broader dataset design can be extended beyond explanation generation. Future work could focus on collecting or generating recovery responses and studying whether models can learn socially appropriate strategies such as clarification, deflection, or apology after a failed humorous attempt.

Finally, the project can be extended to questions of appropriateness. Humor that succeeds is not always socially appropriate, and humor that fails can fail for different reasons. A future dataset could include paired contexts in which the same humorous attempt succeeds in one situation but fails in another, allowing us to study whether models can distinguish between what is funny and what is socially appropriate.

Overall, the current work provides a foundation rather than an endpoint. The lessons learned from prompt development all create opportunities for future research on humor understanding, humor generation, and socially aware language modeling.

\section*{Acknowledgements}

This work was partially supported by NSF IIS-2137396, IIS-2142827, IIS-2234058, and their REU programs. We also appreciate the support from the Foundation Models and Applications Lab of Lucy Institute and ND-IBM Tech Ethics Lab.


\bibliography{custom}

\begin{thebibliography}{22}
\providecommand{\natexlab}[1]{#1}

\bibitem[{Annamoradnejad and Zoghi(2020)}]{annamoradnejad2020colbert}
Issa Annamoradnejad and Gohar Zoghi. 2020.
\newblock Colbert: Using bert sentence embedding for humor detection.
\newblock \emph{arXiv preprint arXiv:2004.12765}, 1(3).

\bibitem[{Baluja(2025)}]{baluja-2025-text}
Ashwin Baluja. 2025.
\newblock \href {https://aclanthology.org/2025.chum-1.2/} {Text is not all you need: Multimodal prompting helps {LLM}s understand humor}.
\newblock In \emph{Proceedings of the 1st Workshop on Computational Humor (CHum)}, pages 9--17, Online. Association for Computational Linguistics.

\bibitem[{Cao et~al.(2025)Cao, Cao, Hou, and Ji}]{CAO2025100807}
Yi~Cao, Jiahao Cao, Yubo Hou, and Li-Jun Ji. 2025.
\newblock \href {https://doi.org/10.1016/j.chbr.2025.100807} {How humorous is ai? exploring chatgpt's role in humor generation and human-ai interaction}.
\newblock \emph{Computers in Human Behavior Reports}, 20:100807.

\bibitem[{Christ et~al.(2025)Christ, Amiriparian, Kathan, Müller, König, and Schuller}]{10707160}
Lukas Christ, Shahin Amiriparian, Alexander Kathan, Niklas Müller, Andreas König, and Björn~W. Schuller. 2025.
\newblock \href {https://doi.org/10.1109/TAFFC.2024.3475736} {Towards multimodal prediction of spontaneous humor: A novel dataset and first results}.
\newblock \emph{IEEE Transactions on Affective Computing}, 16(2):844--860.

\bibitem[{Christ et~al.(2024)Christ, Amiriparian, K\"{o}nig, Eulitz, Cambria, and Schuller}]{10.1145/3689062.3695939}
Lukas Christ, Shahin Amiriparian, Andreas K\"{o}nig, Simone Eulitz, Erik Cambria, and Bj\"{o}rn~W. Schuller. 2024.
\newblock \href {https://doi.org/10.1145/3689062.3695939} {Muse '24: The 5th multimodal sentiment analysis challenge and workshop: Social perception \& humor}.
\newblock In \emph{Proceedings of the 5th on Multimodal Sentiment Analysis Challenge and Workshop: Social Perception and Humor}, MuSe'24, page 10–11, New York, NY, USA. Association for Computing Machinery.

\bibitem[{Gr\'{o}sz et~al.(2023)Gr\'{o}sz, Virkkunen, Porjazovski, and Kurimo}]{10.1145/3606039.3613102}
Tam\'{a}s Gr\'{o}sz, Anja Virkkunen, Dejan Porjazovski, and Mikko Kurimo. 2023.
\newblock \href {https://doi.org/10.1145/3606039.3613102} {Discovering relevant sub-spaces of bert, wav2vec 2.0, electra and vit embeddings for humor and mimicked emotion recognition with integrated gradients}.
\newblock In \emph{Proceedings of the 4th on Multimodal Sentiment Analysis Challenge and Workshop: Mimicked Emotions, Humour and Personalisation}, MuSe '23, page 27–34, New York, NY, USA. Association for Computing Machinery.

\bibitem[{Hyun et~al.(2023)Hyun, Sung-Bin, Han, Yu, and Oh}]{chen2023smile}
Lee Hyun, Kim Sung-Bin, Seungju Han, Youngjae Yu, and Tae-Hyun Oh. 2023.
\newblock Smile: A multimodal dataset for understanding laughter.
\newblock \emph{arXiv preprint arXiv:2312.09818}.

\bibitem[{Joshi(2025)}]{joshi-2025-evaluating}
Narendra~Nath Joshi. 2025.
\newblock \href {https://aclanthology.org/2025.chum-1.9/} {Evaluating human perception and bias in {AI}-generated humor}.
\newblock In \emph{Proceedings of the 1st Workshop on Computational Humor (CHum)}, pages 79--87, Online. Association for Computational Linguistics.

\bibitem[{Kolomaznik et~al.(2024)Kolomaznik, Petrik, Slama, and Jurik}]{kolomaznik2024role}
Michal Kolomaznik, Vladimir Petrik, Michal Slama, and Vojtech Jurik. 2024.
\newblock The role of socio-emotional attributes in enhancing human-ai collaboration.
\newblock \emph{Frontiers in psychology}, 15:1369957.

\bibitem[{Lee et~al.(2025)Lee, Fong, Le, Shah, Han, and Zhu}]{lee-etal-2025-pragmatic}
Joshua Lee, Wyatt Fong, Alexander Le, Sur Shah, Kevin Han, and Kevin Zhu. 2025.
\newblock \href {https://aclanthology.org/2025.chum-1.7/} {Pragmatic metacognitive prompting improves {LLM} performance on sarcasm detection}.
\newblock In \emph{Proceedings of the 1st Workshop on Computational Humor (CHum)}, pages 63--70, Online. Association for Computational Linguistics.

\bibitem[{Lemmens and De~Marez(2026)}]{10.1145/3778357}
Jens Lemmens and Victor De~Marez. 2026.
\newblock \href {https://doi.org/10.1145/3778357} {Computational humor modeling: A survey on the state of the art}.
\newblock \emph{ACM Comput. Surv.}, 58(7).

\bibitem[{Li et~al.(2020)Li, Rzepka, Ptaszynski, and Araki}]{LI2020102290}
Da~Li, Rafal Rzepka, Michal Ptaszynski, and Kenji Araki. 2020.
\newblock \href {https://doi.org/10.1016/j.ipm.2020.102290} {Hemos: A novel deep learning-based fine-grained humor detecting method for sentiment analysis of social media}.
\newblock \emph{Information Processing and Management}, 57(6):102290.

\bibitem[{Marone(2025)}]{marone-2025-algorithm}
Vittorio Marone. 2025.
\newblock \href {https://aclanthology.org/2025.chum-1.11/} {The algorithm is the message: Computing as a humor-generating mode}.
\newblock In \emph{Proceedings of the 1st Workshop on Computational Humor (CHum)}, pages 96--100, Online. Association for Computational Linguistics.

\bibitem[{Martin and Ford(2006)}]{martin2006psychology}
Rod~A Martin and Thomas Ford. 2006.
\newblock The psychology of humor.
\newblock \emph{Burlington, MA: Elsevier}, 2.

\bibitem[{Mirowski et~al.(2025)Mirowski, Mathewson, and Branch}]{mirowski-etal-2025-theate}
Piotr Mirowski, Kory Mathewson, and Boyd Branch. 2025.
\newblock \href {https://aclanthology.org/2025.chum-1.10/} {The theater stage as laboratory: Review of real-time comedy {LLM} systems for live performance}.
\newblock In \emph{Proceedings of the 1st Workshop on Computational Humor (CHum)}, pages 88--95, Online. Association for Computational Linguistics.

\bibitem[{Olafsson et~al.(2020)Olafsson, O'Leary, and Bickmore}]{10.1145/3383652.3423915}
Stefan Olafsson, Teresa~K. O'Leary, and Timothy~W. Bickmore. 2020.
\newblock \href {https://doi.org/10.1145/3383652.3423915} {Motivating health behavior change with humorous virtual agents}.
\newblock In \emph{Proceedings of the 20th ACM International Conference on Intelligent Virtual Agents}, IVA '20, New York, NY, USA. Association for Computing Machinery.

\bibitem[{Sarrof(2025)}]{sarrof-2025-homophonic}
Yash~Raj Sarrof. 2025.
\newblock \href {https://aclanthology.org/2025.chum-1.4/} {Homophonic pun generation in code mixed {H}indi {E}nglish}.
\newblock In \emph{Proceedings of the 1st Workshop on Computational Humor (CHum)}, pages 23--31, Online. Association for Computational Linguistics.

\bibitem[{Singh et~al.(2024)Singh, Kim, Kim, Park, Choi, and De~Neve}]{singh2024exploring}
Ankit~Kumar Singh, Ganghyun Kim, Jaeheon Kim, Ho-min Park, Bong~Jun Choi, and Wesley De~Neve. 2024.
\newblock Exploring multimodal approaches and fusion methods for ceo social attribute prediction in 2024 muse-perception.
\newblock In \emph{Proceedings of the 5th on Multimodal Sentiment Analysis Challenge and Workshop: Social Perception and Humor}, pages 36--44.

\bibitem[{Skalicky and Attardo(2025)}]{skalicky-attardo-2025-testing}
Stephen Skalicky and Salvatore Attardo. 2025.
\newblock \href {https://aclanthology.org/2025.chum-1.6/} {Testing humor theory using word and sentence embeddings}.
\newblock In \emph{Proceedings of the 1st Workshop on Computational Humor (CHum)}, pages 58--62, Online. Association for Computational Linguistics.

\bibitem[{Suls(1972)}]{suls1972twostage}
Jerry Suls. 1972.
\newblock A two-stage model for the appreciation of jokes and cartoons: An information-processing analysis.
\newblock \emph{Humor: International Journal of Humor Research}, pages 81--100.

\bibitem[{Toplyn and Amir(2025)}]{toplyn-amir-2025-ai}
Joe Toplyn and Ori Amir. 2025.
\newblock \href {https://aclanthology.org/2025.chum-1.8/} {Can {AI} make us laugh? comparing jokes generated by witscript and a human expert}.
\newblock In \emph{Proceedings of the 1st Workshop on Computational Humor (CHum)}, pages 71--78, Online. Association for Computational Linguistics.

\bibitem[{Weller and Seppi(2019)}]{weller2019humor}
Orion Weller and Kevin Seppi. 2019.
\newblock Humor detection: A transformer gets the last laugh.
\newblock In \emph{Proceedings of the 2019 Conference on Empirical Methods in Natural Language Processing and the 9th International Joint Conference on Natural Language Processing (EMNLP-IJCNLP)}, pages 3621--3625.

\end{thebibliography}

\appendix
\section{Prompt 1}
\begin{verbatim}
You are annotating humor instances for a 
research dataset.

Your task is to explain why a humorous 
attempt succeeds or fails in context.

Input fields:
- C_text: preceding context or dialogue
- C_completeness: one of {complete, 
incomplete, minimal}
- X_text: the utterance that may be 
humorous
- X_speaker: speaker of X_text
- Y: receiver reaction, where 1 = humor 
succeeded / audience laughed, 0 = humor 
failed
/ no laughter

Output requirements:
Return exactly one JSON object with 
this structure:

{
"reasoning_steps": [
"Step 1: ...",
"Step 2: ...",
"Step 3: ..."
],
"flow_chart": [
"...",
"...",
"..."
],
"final_explanation": "...",
"hallucination_flag": 0
}

Instructions:
1. Explain the humor as a short, explicit, 
step-by-step reasoning process.
2. Ground every step in the provided 
context whenever possible.
3. If commonsense, social expectations, or 
cultural knowledge is needed, explicitly 
say so using phrases like:
- "Commonsense knowledge: ..."
- "People generally expect ..."
- "This suggests ..."
4. Do not invent specific facts not 
supported by the input.
5. If the context is incomplete, say what 
is missing and make only minimal, cautious 
inferences.
6. The flow_chart should be a causal 
sequence showing how the context builds 
expectation and how X_text changes it. 
Someone should be able to recreate the 
joke from the flow chart. 
8. final_explanation must be 1 sentence.
9. Set hallucination_flag to:
- 0 if the explanation is well-supported
by the given text and commonsense
- 1 if the explanation depends on 
uncertain assumptions or missing context
10. Keep the output concise and readable.

\end{verbatim}

\section{Prompt 2}
\begin{verbatim}
You are annotating humor instances for a 
research dataset.

Your task is to explain why a humorous 
attempt succeeds or fails in context, 
while also checking whether the dataset 
instance itself may be flawed.

Use only the text you are given, plus 
ordinary commonsense and very broadly 
shared public cultural knowledge when 
that knowledge is directly cued by the 
text.
Do not use previous chat history.
Do not assume the dataset is correct.
Do not force a humor explanation when 
the textual evidence is weak.

Input fields:
- C_text: preceding context or dialogue
- X_text: utterance that may be humorous
- X_speaker: speaker of X_text
- Y: receiver reaction, where 1 = humor 
succeeded / audience laughed, 0 = humor 
failed / no laughter

Output requirements:
Return exactly one JSON object with this 
structure:

{
  "text_support_label": "text_sufficient 
 | text_insufficient_possible_multimodal |
 text_insufficient_missing_context |
 text_insufficient_possible_transcription
 _error",
  "data_quality_flags": [
    "missing_context | 
    possible_bad_transcription |
    possible_speaker_misattribution | 
    possible_truncation | 
    possible_alignment_error |
    possible_multimodal_dependence"
  ],
  "reasoning_steps": [
    "Step 1: ...",
    "Step 2: ...",
    "Step 3: ..."
  ],
  "flow_chart": [
    "...",
    "...",
    "..."
  ],
  "final_explanation": "...",
  "hallucination_flag": 0
}

Core decision rule:
Mark an instance as "text_sufficient" 
ONLY if all of the following are recoverable 
from C_text + X_text + ordinary broadly 
shared knowledge directly cued by the text:
1. the exact humor trigger phrase or idea,
2. the relevant reference / association 
/ comparison behind that trigger,
3. how that association maps onto the local
context,
4. why that mapping would plausibly produce
laughter here.

If any of those 4 pieces is missing, do NOT
mark it as text_sufficient.

Instructions:
1. First decide whether the humor reaction 
is explainable from the text alone.
2. Then decide whether the transcript 
appears clean or flawed.
3. Treat Y=1 as evidence that the audience 
laughed, NOT as evidence that the text is 
sufficient.
4. Use 
"text_insufficient_possible_multimodal" 
when the laughter likely depends on an 
unseen slide, image, gesture, facial 
expression, timing, delivery, or other 
live visual/performance cue.
5. Use 
"text_insufficient_missing_context" when 
the joke seems to require omitted prior 
dialogue, background, or setup that is 
not present in the text.
6. Use 
"text_insufficient_possible
_transcription_error" when the text 
appears malformed, mistranscribed, 
semantically broken, oddly segmented, or
otherwise unreliable.
7. Add one or more data_quality_flags 
whenever there are concrete textual 
signs of a flawed instance.
8. Do not invent missing content, 
missing visuals, repaired wording, or 
hidden setup.
9. If commonsense or public cultural 
knowledge is needed, explicitly name it.
10. If a named entity, title, slogan, 
song, proverb, thought experiment, 
nickname, or quotation is central to the
joke, explain that reference briefly.
11. However, do NOT assume niche or 
specialized knowledge. If the humor only
works with non-obvious historical, 
domain-specific, or poorly cued 
knowledge, prefer an insufficient label.
12. If X_text contains demonstratives or
unresolved references such as "this", 
"that", "here", "there", "like this", 
"this version", "look", or "something 
like this", and the referent is not 
clearly recoverable from the text, do 
NOT mark text_sufficient.
13. When such unresolved language 
likely points to something being 
shown or performed live, prefer
"text_insufficient_possible_multimodal".
14. Do not accept generic explanations 
such as "clever", "ironic", "unexpected 
twist", "funny name", or "play on words"
unless you explicitly identify:
    - what the exact trigger is,
    - what it refers to,
    - and how that creates humor here.
15. If you cannot explain why a word, 
name, label, quotation, or comparison 
is funny, apt, or meaningful, do NOT 
mark text_sufficient.
16. If the actual humor trigger may 
occur earlier in C_text and X_text is 
only a reaction, continuation, or 
callback, say so explicitly. If the 
laughter may be aligned to an earlier 
line, consider possible_alignment_error.
17. Ground every reasoning step in the 
provided text.
18. reasoning_steps must do exactly 
these 3 things:
    - Step 1: identify the likely humor 
    trigger phrase or idea,
    - Step 2: explain the relevant 
    commonsense / cultural / semantic 
    association behind it,
    - Step 3: explain how that 
    association fits this local context
    OR explain exactly why the 
    transcript is insufficient to do 
    so.
19. The flow_chart must be 
reconstructive, not generic. It should 
let a reader either:
    - recreate the joke mechanism from 
    the text alone, OR
    - understand precisely why the 
    transcript is insufficient.
20. The flow_chart should usually have 4 to 
6 short steps, not filler.
21. In the flow_chart, explicitly include:
    - the setup or expectation,
    - the trigger phrase,
    - the intended association/reference,
    - the shift/incongruity,
    - and, if applicable, the missing link 
    or unresolved referent.
22. If the instance is not text_sufficient,
the flow_chart must explicitly say what is 
missing or unresolved.
23. final_explanation must be exactly 1 
sentence.
24. Set hallucination_flag to:
   - 0 if the conclusion is well-supported 
   by the given text and broadly shared 
   knowledge directly cued by the text
   - 1 if the conclusion depends on 
   uncertain assumptions, unresolved 
   references, suspected multimodality, 
   missing context, or suspected transcript 
   flaws
25. Keep the output concise and readable.

Important distinctions:
- Broadly shared public knowledge that is 
directly cued by the text IS allowed.
  Example types: a famous song title, 
  a very well-known movie reference, or 
  Schrödinger's cat if the text directly 
  cues it.
- But if the joke still depends on an 
unresolved referent, hidden slide, 
unknown quote source, niche historical 
saying, or unexplained name/label, do NOT 
mark text_sufficient.
- A correct explanation must be specific 
enough that another reader can see exactly 
why the line is funny, not just that it is 
"clever" or "unexpected."
\end{verbatim}

\end{document}